\documentclass[preprint,12pt]{elsarticle}

\usepackage{amssymb}
\usepackage{url,hyperref,lineno,microtype,subcaption}
\usepackage[onehalfspacing]{setspace}
\usepackage{caption}
\usepackage{booktabs}
\usepackage{multirow,booktabs}
\usepackage{amsmath} 
\usepackage{makecell}
\usepackage{hyperref} 

\begin{document}

\begin{frontmatter}



\title{A Brain-inspired Computational Model for Human-like Concept Learning}


\author[address1,address2,address3]{Yuwei Wang}
\author[address1,address2,address3]{Yi Zeng\corref{cor1}}
\cortext[cor1]{Corresponding author at: Institute of Automation, Chinese Academy of Sciences, Beijing, 100190, China}
\address[address1]{Brain-inspired Cognitive Intelligence Lab, Institute of Automation, Chinese Academy of Sciences, Beijing, 100190, China.}
\address[address2]{International Research Center for AI Ethics and Governance, Institute of Automation, Chinese Academy of Sciences, Beijing, 100190, China.}
\address[address3]{Center for Long-term Artificial Intelligence, Beijing, China.}

\begin{abstract}
Concept learning is a fundamental part of human cognition and plays a crucial role in mental processes such as categorization, reasoning, memory and decision making.
Researchers from different fields have consistently been interested in the issue of how people acquire concepts.
To clarify the mechanism of concept learning in humans, this paper reviews the findings in computational neuroscience and cognitive psychology. 
They reveal that multisensory representation and text-derived representation are the two essential components of the brain's representation of concepts. 
The two types of representations are coordinated by a semantic control system, and concepts are eventually learned.
Inspired by this mechanism, this paper constructs a human-like computational model for concept learning based on spiking neural networks.
The dilemmas of diverse sources and imbalanced dimensionality of the two forms of concept representations are effectively overcome to obtain human-like concept representations. 
Similar concepts tests reveal that our model, which is created in the same way that humans learn concepts, provides representations that are also more similar to human cognition.

\end{abstract}




\end{frontmatter}






\bibliographystyle{elsarticle-num-names} 


\section{Introduction}
Concept acquisition is an important foundation for many human cognitive tasks.
Concepts can be learned in different ways. 
For instance, the sound of snow falling, the color of snow, the microscopic shape, the touch, and even the taste of snowflakes can all help one understand the concept of a “snowflake”. 
People who live in low latitudes and low altitude regions and have never experienced snow may still be able to understand the concept through textual descriptions such as \emph{“little snowflake is falling from the sky”}, \emph{“snowflakes differ from each other though they follow similar patterns”} and \emph{“snowflakes form when water vapor condenses around specks of dust high in the clouds”}.

Humans learn concepts approximately in two ways: by interacting with the environment, integrating information from different perceptual modalities, and through textual information, combining the contextual content, grammatical structure and dependency structure of concepts.
Paivio\cite{DualCodeTheory} proposed the dual code theory, which holds that there are two types of cognitive phenomena handled by distinct subsystems: verbal and nonverbal. One has an emphasis on dealing with language, whereas the other has a specialization on representing and analyzing information about nonverbal objects and events.
Barsalou\cite{LASS} proposed the language and situated simulation theory, which takes into account that there are two sources of knowledge: situated simulations in the brain's modal systems and linguistic forms in the language systems of the brain.

Bi \emph{et al.}\cite{BiTwoForms1}\cite{BiTwoForms2} validated this result through experiments in cognitive psychology.
Congenitally blind and normal volunteers rated the color similarity of the same object pair in the behavioral level, showing that the color knowledge spaces of them are remarkably comparable.
According to fMRI testing and analysis, both of them represent object color in the left dorsal anterior temporal lobe, but only normal people do so in the ventral occipitotemporal color perception area, which congenitally blind individuals lack.
The two sets of tests put together revealed the existence of multisensory and linguistic representations in the brain.

Regarding the relationship between multisensory and text-derived representations of concepts in the brain, Giacomo \emph{et al.}\cite{BrainDistributionRepresentation} insist that concepts are distributed, modality-independent cortical representations in the brain. 
They did this by analyzing research on speech production behavior in sighted and congenitally blind individuals as well as fMRI experiments.
Fernandino \emph{et al.}\cite{SeosoryMotorEncodesWordMeaning} came to the observation that the semantic information of concepts is at least largely based on multisensory information such as motion using fMRI data from trials with multisensory input in a semantic decision-making task.
The tri-network concept processing system, or one with three submodules: multimodal experience system, language-supported system, and semantic control system, was proposed by Xu \emph{et al.}\cite{BiTriNetworkImage}\cite{BiTriNetwork} after reviewing earlier related work by analyzing the strength of functional connectivity among various brain regions in the resting state using a graph theory approach.
This allows us to distill a key principle for human concept learning: the brain has two key components for concept representation, one based on text-derived representation and the other on multisensory integration, both of which are coordinated by a semantic control system.
This rule will hopefully serve as a crucial foundation for designing and building a computational model that can learn concepts like human beings.

For the conventional computational models of concept learning, the cue combination model, based on Bayesian decision theory, is a well-known model for examining multisensory integration concept learning in the field of cognitive psychology\cite{Bayes-Nature2002}\cite{Bayes-MCD}\cite{Bayes-BasisFunction1}\cite{Bayes-BasisFunction2}.
It primarily uses the stimuli from various modalities as the input of the likelihood function, integrating the data from different modalities with a weighted linear model.
These weights are determined by making the posterior distribution the maximum under the conditional independence assumption and the Bayes' rule.
Through cognitive-behavioral studies, the best solutions of this class of models have been shown to process information in a manner that is remarkably comparable to that of humans\cite{Bayes-Nature2002}.
However, due to the non-uniformity of the data input format, it is challenging to apply the same procedure to other species for validation, and it is even more challenging to do so when designing computational models.

Researchers in the field of artificial intelligence prefer model design for concept learning with traditional machine learning methods, which consider various modal information as model inputs.
They mostly employ algorithms: 
cancatenate\cite{Cancatenate1}\cite{Cancatenate2}\cite{Cancatenate3}\cite{Cancatenate4},
canonical correlation analysis\cite{CCA}\cite{kCCA},
matrix decomposition\cite{SVD}\cite{MCANMF},
language model with multisensory conctext\cite{ContextMultisensory},
autoencoders\cite{AE1}\cite{AE2}, and 
tensor fusion networks\cite{TFN1}\cite{TFN2}\cite{TFN3}.
These algorithms have achieved good results on tasks like similar concepts and sentiment analysis.
However, the inputs to the models are mainly text, images and speech obtained by perceptrons, ignoring the comprehensiveness of human cognitive behaviors and taking less into account the modalities of touch, smell and taste.
In their algorithms, text information is viewed equally with various other sensory information without distinction, ignoring the unique contribution of text information in human concept learning. 
Knowledge acquired through language is not modeled separately from other multisensory information.
This is inconsistent with the way concepts are represented in the brain and is not very biologically interpretable.

\begin{figure}[h!]
  \centering
  \includegraphics[width=1.0\textwidth]{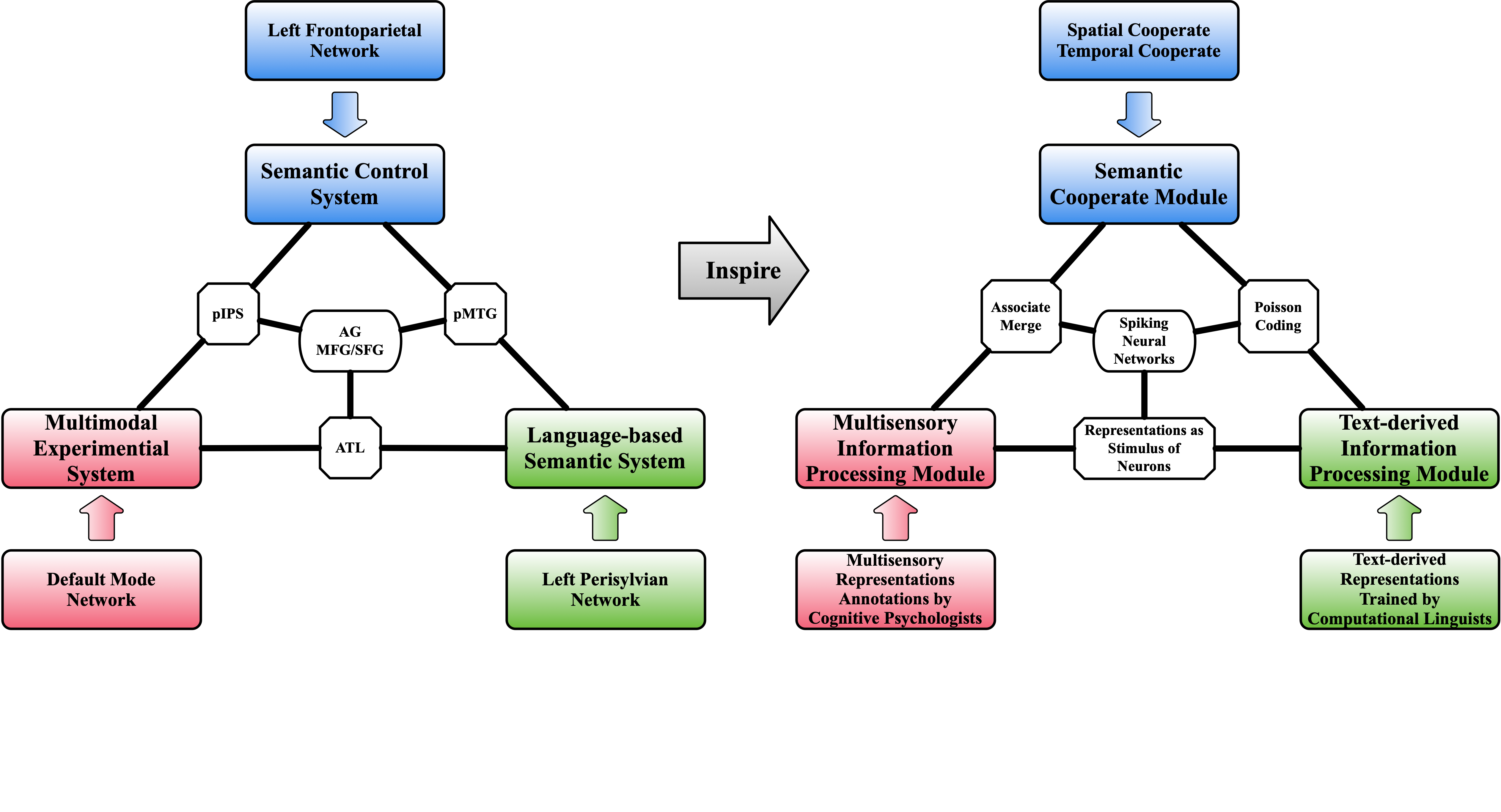}\\
  \caption{The Framework of the Brain-inspired Computational Model on Human-like Concept Learning}
  \label{Inspiration}
\end{figure}

To create concept representations that are more akin to those of humans, we will build a computational model in this research that 
is inspired by human-like mechanism.
By utilizing the spiking neural networks, we cooperate the text-derived and multisensory representations of concepts in a human-like manner.
The complete framework is depicted in Figure \ref{Inspiration}.

\section{Results}
In this study, we develop a computational model for concept learning with spike neural networks in a manner similar to that of humans.
The model's inputs are multisensory representations labeled by cognitive psychologists and text-derived representations developed by computational linguists.
They are correlated to the multimodal experimential system and the language-based semantic system in the brain, respectively.
These two types of concept representation datasets will be introduced first in this section.
The model is fed with the the two types of representations, and the simulation of the semantic control system is completed by the transformation of the spike distribution matrices and the cooperation of spatial-temporal information to generate the final human-like concept representations.
Through the similar concepts task in this section, we will firstly evaluate how well the gennerated concept representations perform.
A case study will provide a more concrete illustration of the advancement of our human-like concept representations and we will demonstrate the parametric analysis in the end.

\subsection{Two Types of Concept Representation Datasets}
The quantitative concept representation datasets highlight two points of view: embodied theory-based multisensory representations and distributed hypothesis-based text-derived representations\cite{AboutConcept1}.
Multisensory representations emphasize that the meaning of concepts is grounded in our sensory experiences, perceptions, movements, and interactions with the environment\cite{Barsalou}. 
Each dimension of such representations represents a sensory modality, and the value represents the intensity with which the concept is acquired through that modality. 
While text-derived representations rely on the distributional hypothesis that the similarity between two concepts is rooted in their contexts\cite{Harris}.
The more similar the context in a statistical sense, the more similar the text-derived representations of the concepts are through the training of a large-scale corpus.

The existing publicly available multisensory representation datasets primarily annotated by cognitive psychologists.
In order to achieve them, participants are asked how much they perceive a concept through various sensory modalities (such as auditory, gustatory, haptic, olfactory, visual, etc.).
The pioneering works of multisensory datasets are the “modality exclusivity norms” which are proposed by Lynott and Connell (423 adjective concepts\footnote{\url{https://link.springer.com/article/10.3758/BRM.41.2.558}}\cite{LC423a} and 400 nominal concepts\footnote{\url{https://link.springer.com/article/10.3758/s13428-012-0267-0}}\cite{LC400n}).
As shown in Table \ref{LC823DEMO}, each concept is characterized by a 5-dimensional vector indicating the strength of association of the concept with each of the five main perceptual modalities.
In this paper, when using them, the two datasets are merged together and denoted as LC823.
Based on more thorough neurobiological evidence to design more specialized modal dimensions, Binder \emph{et al.}\cite{BBSR} constructed brain-based componential semantic representation(BBSR), including 65 dimensions of perceptual, motor, spatial, temporal, emotional, social and cognitive experiences.
This dataset, which has 535 concepts, performs well at capturing semantic similarity and identifying concept categories.
It is the most interpretable one in the current multisensory representation datasets.

\begin{table}[h!]
\begin{center}
\caption{Multisensory Representations of Concepts in LC823}
\label{LC823DEMO}
\scalebox{0.9}{
\begin{tabular}{l|ccccc}
\toprule
Concept	&Auditory	&Gustatory	&Haptic	&Olfactory	&Visual\\
\midrule
academy	&1.411764706	&0.058823529	&0.470588235	&0.117647059	&2.705882353\\
air	&1.058823529	&1.470588235	&2.117647059	&2.529411765	&1.352941176\\
food	&1.470588235	&4.882352941	&3.411764706	&4.529411765	&4.294117647\\
doctor	&3	&0.705882353	&1.647058824	&1.294117647	&4.470588235\\
eye	&0.058823529	&0	&2.235294118	&0	&4.764705882\\
honey	&0.647058824	&4.764705882	&2.823529412	&3.764705882	&4.117647059\\
money	&1.470588235	&0.352941176	&3.705882353	&1.176470588	&4.823529412\\
laughter	&4.882352941	&0	&0.941176471	&0	&4.117647059\\
blanket	&0.352941176	&0.176470588	&4.235294118	&1.647058824	&4\\
file	&0.764705882	&0.117647059	&2.411764706	&0.235294118	&4.058823529\\	
\bottomrule
\end{tabular}
}
\end{center}
\end{table}

Whereas text-derived representations of concepts are mainly obtained by computational linguists through large-scale corpus training.
These methods, which transform semantic and syntactic information of concepts into dense vectors for representation, have been widely used along with the development of natural language processing techniques.
Word2vec\footnote{\url{https://code.google.com/archive/p/word2vec/}} \cite{w2v} and GloVe\footnote{\url{https://nlp.stanford.edu/ projects/glove/}} \cite{Glove} are representative algorithms for obtaining text-derived representations.
Word2vec includes two types of models: continuous bag of words, which learns to predict the current word for a given context, and skip-gram models, which learn to predict the contextual word for a given current word.
GloVe is a specific weighted least squares model that is trained on a word-word co-occurrence count matrix that integrates global matrix decomposition and local contextual information.
The large scale of the corpus ensures their representational power, but the information represented by each dimension of the vector is ambiguous.
The size of such representation datasets is proportional to the number of words in the corpus, and the most widely used public versions are in million-level.

Previous research has shown that both types of representations for concepts, despite coming from distinct sources, are capable of reflecting human cognition from a macroscopic perspective.
While the representation similarity analysis shows that there is a significant gap between the two types of representations, and combining the two can improve the representation effect\cite{wang2022statistical}.
This creates a solid foundation for us to build a computational model for concept learning in a way that is similar to how humans learn, and it is practical and reasonable to utilize them as the computational model's input.

\subsection{Similar Concepts Test}

The similar concepts test is a traditional method for assessing how closely a system or algorithm-generated representation matches human cognition.
This method will also be used in this article.
By utilizing evaluation datasets, we compare the closeness of our model-generated concept representations and several traditional representations to human cognition.

\subsubsection{Performance Metric}
The majority of evaluation datasets used for similar concepts tests are rated by volunteers, and the labelled objects are a variety of concept pairs.
The concept pairs in the evaluation dataset MEN are shown in Table \ref{MEN} together with the participants' corresponding similarity scores.
SimLex999, MEN, and MTurk771 are the three evaluation datasets that will be used in this study, and a brief description of them is provided below.

\begin{table}[h!]
\begin{center}
\caption{The Demonstration of Similar Concept Dataset(MEN)}
\label{MEN}
\scalebox{0.75}{
\begin{tabular}{cc||cc||cc}
\toprule
Concept Pairs&Ratings&Concept Pairs&Ratings&Concept Pairs&Ratings\\
\midrule
\emph{(automobile,\ car)}&50&\emph{(grass,\ lawn)}&48&\emph{(children,\ kids)}&46\\
\emph{(bread,\ sandwich)}&46&\emph{(ice,\ snow)}&46&\emph{(coffee,\ tea)}&45\\
\emph{(snow,\ winter)}&44&\emph{(feathers,\ peacock)}&43&\emph{(camera,\ lens)}&43\\
\emph{(rain,\ wet)}&42&\emph{(pizza,\ restaurant)}&39&\emph{(office,\ table)}&34\\
\emph{(harbor,\ sea)}&38&\emph{(brick,\ concrete)}&37&\emph{(cup,\ town)}&39\\
\emph{(cherry,\ orange)}&34&\emph{(black,\ yellow)}&32&\emph{(cold,\ wet)}&31\\
\emph{(cliff,\ waterfall)}&30&\emph{(hot,\ wet)}&29&\emph{(cute,\ happy)}&27\\
\emph{(panda,\ pigs)}&26&\emph{(hill,\ river)}&23&\emph{(evening,\ sunshine)}&22\\
\emph{(autumn,\ frost)}&22&\emph{(porch,\ tower)}&20&\emph{(lake,\ town)}&17\\
\emph{(frozen,\ soup)}&17&\emph{(blue,\ happy)}&16&\emph{(bacon,\ sweet)}&16\\
\emph{(dirty,\ fun)}&15&\emph{(apartment,\ valley)}&14&\emph{(paper,\ vine)}&14\\
\emph{(nature,\ tv)}&14&\emph{(book,\ building)}&14&\emph{(breakfast,\ floor)}&12\\
\emph{(eat,\ shade)}&11&\emph{(cute,\ dirty)}&11&\emph{(coast,\ zebra)}&10\\
\emph{(hot,\ winter)}&9&\emph{(banana,\ feet)}&9&\emph{(desert,\ roof)}&8\\
\emph{(ski,\ skirt)}&7&\emph{(dragon,\ oak)}&7&\emph{(snake,\ taxi)}&7\\
\emph{(cow,\ table)}&6&\emph{(museum,\ swim)}&6&\emph{(jump,\ salad)}&6\\
\emph{(monkeys,\ restaurant)}&5&\emph{(hill,\ rust)}&5&\emph{(brick,\ rabbit)}&5\\

\emph{(meat,\ pond)}&5&\emph{(gun,\ pizza)}&4&\emph{(fish,\ theatre)}&3\\
\emph{(cafe,\ lizard)}&2&\emph{(giraffe,\ harbor)}&1&\emph{(bakery,\ zebra)}&0\\

\bottomrule
\end{tabular}
}
\end{center}
\end{table}

SimLex999 is a similar concepts evaluation dataset proposed by Hill \emph{et al.}\cite{SimLex999} that puts more emphasis on inter-concept similarity and explicitly quantifies the similarity between concepts by human annotation.
The dataset, which includes 1,028 concepts and 999 concept pairs, was made available to the public in  \url{https://fh295.github.io/simlex.html} and was annotated by 500 volunteers recruited using Amazon Mechanical Turk.
The MTurk771 similar concepts dataset, which is described by Halawi \emph{et al.} \cite{MTurk771},  was annotated by recruiting volunteers through Amazon Mechanical Turk. 
It focuses on both the similarity and the association between concepts.
The low quality annotations are avoided by creating an experimental scheme with statistical techniques, and the mean value is determined as the concept pairs' similarity score.
The data is freely accessible at \url{http://www2.mta.ac.il/~gideon/datasets/} and consists of 771 concept pairs and 1,113 individual concepts.
Bruni \emph{et al.} 's MEN\cite{SVD} is a dataset that also used Amazon Mechanical Turk to gather volunteers for annotation.
In contrast to the previous two datasets, the annotation strategy is different.
Volunteers are asked to determine which set of concept pairs is more actually similar in two pairs rather than assigning each concept pair a direct numerical score.
The final step was to determine the similarity scores of the two concepts in the concept pair by normalizing the data.
The dataset, which is publically accessible at \url{http://clic.cimec.unitn.it/~elia.bruni/MEN}, has a total of 751 concepts and 3,000 concept pairs.

The three evaluation datasets mentioned above are all quantitative descriptions of human-acquired concepts, and the level of scoring represents the degree of conceptual similarity.
They will be used as the evaluation criteria of human cognition in this paper.

How to quantify the relationship between human cognition and the concept representations generated by the computational model?
Due to the fact that concept pairs' similarity scores differ depending on the evaluation datasets, it is crucial to consider the concept pairs' similarity rankings within these datasets rather than the similarity scores.
The Spearman's correlation coefficient between similar rankings of the same concept pair in the evaluation dataset and in the representation space is the most popular approach to quantify the closeness between the computational model representation space and the human cognitive space.
As indicated in Table \ref{SpearmanComputationDEMO}, all concept pairs in the evaluation dataset are ranked based on annotation ratings, and the corresponding rankings are determined.
Simultaneously, similarity is computed and ranked for the same concept pair in the representation space using the distance metric function (cosine or Hamming distance), and the associated rankings are obtained.
Finally, as a measure of closeness, Spearman's correlation coefficient is calculated for the two sets of rankings.

\begin{table}[h!]
\begin{center}
\caption{The Method to Calculate Closeness}
\label{SpearmanComputationDEMO}
\scalebox{0.80}{
\begin{tabular}{c|cc|cc}
\toprule
Concept Pair&\makecell[c]{Ratings \\ by Human}&\makecell[c]{Rankings\\of Human Annotated \\ Ratings}&\makecell[c]{Similarity in \\Representation \\ Datasets}&\makecell[c]{Rankings in \\Representation \\ Datasets}\\
\midrule
\emph{(tiger,\ tiger)}&10&1&1&1\\
\emph{(computer,\ keyboard)}&7.62&5&0.987587052&3\\
\emph{(plane,\ car)}&5.77&9&0.967001033&5\\
\emph{(train,\ car)}&6.31&7&0.969252691&4\\
\emph{(football,\ soccer)}&9.03&2&0.900853568&7\\
\emph{(law,\ lawyer)}&8.38&3&0.735108495&11\\
...&...&...&...&...\\
...&...&...&...&...\\
\emph{(tiger,\ zoo)}&5.87&8&0.840800623&9\\
\emph{(minister,\ party)}&6.63&6&0.918485946&6\\
\emph{(problem,\ airport)}&2.38&11&0.755135699&10\\
\emph{(day,\ summer)}&3.94&10&0.863629918&8\\
\emph{(man,\ woman)}&8.3&4&0.997714718&2\\

\bottomrule
\end{tabular}
}
\end{center}
\end{table}

\subsubsection{Results and Analysis}
In this research, two typical datasets will be employed in both multisensory(LC823 and BBSR) and text-derived(word2vec and GloVe) representations, while three similar concept evaluation datasets(SimLex999, MEN, and MTurk771) will be used to test the computational model.
The comparison trials in the similar concepts test are the multisensory representation or text-derived representation of the concept alone, and the representation in which the two representations of the concept are directly concatenated together, as is commonly used in traditional practice.

Given that the computational model used to learn concepts is an unsupervised learning model, only the outcomes of the generated representations can be used to test the model's effectiveness.
In order to pick the optimal results for analysis, the parameters of spatial stride and temporal stride are explored in this study.
In all experiments, the recording interval $T$ of neural activity is uniformly set to 1000. 
The range of the spatial and temporal stride traversals is $[1, 2, 3, 4, 5, 6, 7, 8, 9, 10, 20, 30, 40, 50, 60, 70, 80, 90, 100]$ and the results demonstrated will automatically filter out the results with diversity less than or equal to 0.05.

In Table \ref{SimilarConceptsResults}, the experimental findings are displayed.
On all three evaluation datasets, the computational model proposed in this research that combines the multisensory information of concepts with information extracted from text in a human-like manner yields better results.
This method effectively combines data from two sources, and the model's cooperation in both the spatio-temporal dimensions increases its representational capacity.
It is superior to the uncoordinated concept representations and also more reliable and cognition-like than the direct concatenation method.
In contrast, the traditional direct concatenation approach is not necessarily better, but even worse.
The two forms of concept representation are not well reconciled by the direct concatenation approach, possibly due to an imbalance in the characteristics of the two types of representations.
Additionally,  the concatenation approach also exhibits bias in similarity calculations and does not guarantee a stable representation effect.
The spiking neural networks in our model effectively avoids this imbalance and improves the generated human-like concept representations.

\begin{table}[h!]
\begin{center}
\caption{The Results of  Human-like Similar Concepts Tests}
\label{SimilarConceptsResults}
\begin{tabular}{lcccc}
\toprule
Combinations & Types	& SimLex999 &MEN  &MTurk771\\	
 \midrule
\multirow{4}*{w2v-LC823} &  Text	&0.203760684	&0.648088252	&0.442105263\\
&Multisensory	&0.348376068	&0.296877883	&0.515789474\\
&Concatenate	&0.298119658	&0.341074126	&0.493233083	\\
&COO	&\underline{0.641880342}	&\underline{0.851279567}	&\underline{0.803007519}\\
 \midrule
\multirow{4}*{w2v-BBSR} &Text &0.214303752 &0.576522874 &0.621428571\\
&Multisensory	&0.355862193	&0.41111554	&0.360714286\\
&Concatenate	&0.27245671	&0.58290295	&0.492857143\\
&COO	&\underline{0.554292929}	&\underline{0.957985015}	&\underline{0.714285714}\\
  \midrule
 \multirow{4}*{GloVe-LC823} &Text	 &0.46017094	&0.731785457	&0.357894737\\
&Multisensory	&0.348376068	&0.321806294	&0.515789474\\
&Concatenate	&0.337094017	&0.358885876	&0.464661654\\
&COO	&\underline{0.687350427}	&\underline{0.855246173}	&\underline{0.781954887}\\
  \midrule
 \multirow{4}*{GloVe-BBSR} &Text	&0.329274892	&0.74939797	&0.721428571\\
&Multisensory	&0.355862193	&0.41111554	&0.360714286\\
&Concatenate	&0.379166667	&0.70415785	&0.657142857\\
&COO	&\underline{0.544608422	}&\underline{0.954678276}	&\underline{0.775}\\
\bottomrule
 
\end{tabular}
\end{center}
\end{table}

\subsection{Case Study}
This part will concentrate on providing the outcomes in the form of case studies, in contrast to the above quantitative demonstration of the effects of the proposed computational model for human-like concept learning using macro metrics.
We will go into detail about the three concept representations used in the evaluation dataset MTurk771, including multisensory representations, text-derived representations, and human-like concept representations based on our computational model.
As shown in Table \ref{CaseStudy}, it primarily contains the scores and corresponding rankings of each concept pair. 
These rankings are based on the multisensory representations alone, the text-derived representations alone, and our human-like concept representations.
In terms of macroscopic closeness metrics, our human-like conceptual representation is closer to human beings.
From a microscopic standpoint, the rankings for the top five concept pairs ranked by human cognition are 10, 3, 7, 13, and 8, and those based solely on multisensory representations are 3, 9, 6, 13, and 8. However, the rankings of human-like concept representations are significantly better, coming in at 6, 1, 9, 2, and 7.
The rankings for the bottom five conceptual pairs of human cognition are 20, 6, 13, 12, and 19 for text-based representations alone, 11, 10, 12, and 20, and 17 for multisensory representations alone. 
For the human-like conceptual representations, the rankings are 11, 15, 17, 19, and 20, which are also noticeably better.
More noteworthy are the concepts “blue” and “red”, where both the original multisensory and text-derived representations are listed as the most analogous concepts, yet this is contrary to how people typically perceive these colors.
The human annotation rating in MTurk771 is ranked 10$th$, while our human-like representation is ranked 14$th$, demonstrating that our model can synergize the two types of representations well and bring the generated representations closer to humans.

\begin{table}[h!]
\begin{center}
\caption{The Case Study of Similar Concepts Test}
\label{CaseStudy}
\scalebox{0.75}{
\begin{tabular}{ll|cc|ccc}
\toprule
Concept1	&Concept2	&\makecell[c]{Rating \\ in MTurk771}	 &\makecell[c]{Ranking \\ in MTurk771}	 &\makecell[c]{Ranking Based on\\ Text-derived \\ Representations}	&\makecell[c]{Ranking Based on \\ Multisensory \\ Representations}	&\makecell[c]{Ranking in \\Human-like \\ Representations}\\
   \midrule
pupil	&student	&4.52&1	&10	&3	&6 \\
aim	&purpose	&4.36&2	&3	&9	&1 \\
cousin	&relation	&4.04&3	&7	&6	&9\\
form	&type	&3.91&4	&2	&13	&2\\
hole	&opening	&3.76&5	&14	&8	&7\\
account	&statement	&3.68&6	&8	&2	&3\\
health	&welfare	&3.5&7	&4	&7	&4\\
mate	&relation	&3.43&8	&15	&4	&12\\
mouth	&opening	&3.30&9 &17	&18	&10\\
blue	&red	&3.27&10	&1	&1	&14\\
matter	&text	&3.27&11	&11	&19	&8\\
construction	&window	&2.76&12	&18	&15	&13\\
call	&meeting	&2.73&13	&5	&14	&16\\
minute	&quantity	&2.61&14	&16	&16	&5\\
measure	&money	&2.57&15	&9	&5	&18\\
plane	&tool	&2.30&16	&20	&11	&11\\
call	&statement	&2.13&17	&6	&10	&15\\
amount	&distance	&1.96&18	&13	&12	&17\\
knowledge	&taste	&1.87&19	&12	&20	&20\\
foot	&recognition	&1.43&20	&19	&17	&19\\
\bottomrule
\end{tabular}
}
\end{center}
\end{table}

\subsection{Parameters Analysis}
The spatial stride $ss$, temporal stride $ts$, and the way used to execute the cooperation of the two types of representations are the three essential parameters in our computational model for human-like concept learning.
We will analyze these parameters in this section.
The following quantitative expression can be used in the model to express the relationship between the spatial step $ss$, the temporal step $ts$, and the ultimate output dimensions provided by this model.

\begin{equation}
Output \ Dimensions = \lceil \frac{D_{Text}-D_{ms}^{AM}}{ss} \rceil  * \lceil \frac{D_{ms}^{AM} * T}{ts} \rceil
\end{equation}

\begin{figure}[h!]
  \centering
  \includegraphics[width=0.9\textwidth]{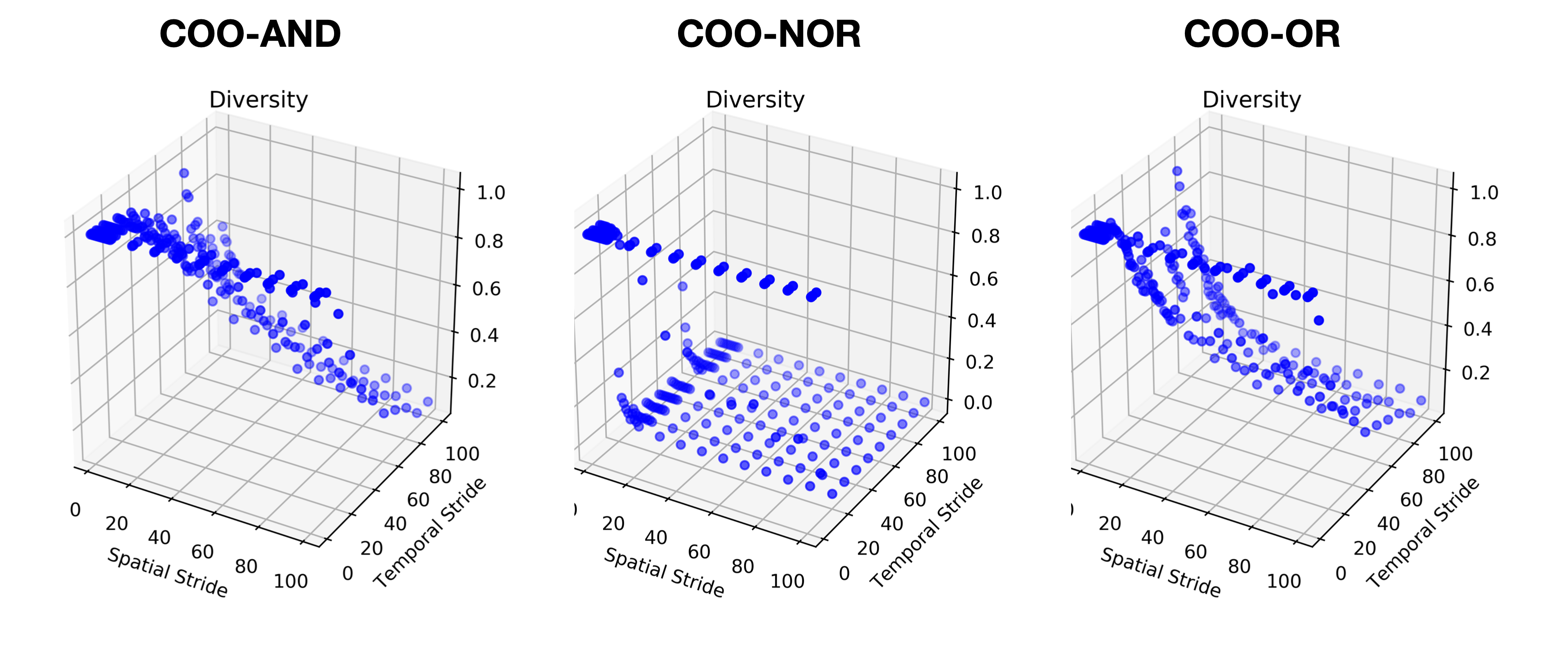}\\
  \caption{The Relation Between Representation Diversity and Spatial Stride, Temporal Stride}
  \label{DiversityAM}
\end{figure}

This indicates that the final human-like concept representations' dimensionality directly depend on the size of the two strides in the model, and this impact is accompanied by differences in representational power and storage space.
We visualize the relationship between the spatial step size $ss$, the temporal step size $ts$, and the representational diversity by defining the representational diversity, i.e., the ratio of the number of all different representations of a concept to the number of all concepts, in order to further quantify the impact of these two parameters on the representational capacity.
Regardless of the cooperation strategy utilized, Figure \ref{DiversityAM} clearly illustrates how the diversity of representations reduces as the spatial and temporal strides increase.
Smaller spatial and temporal strides also imply that the generated representations have a higher dimensionality and will require more storage capacity.
In order to achieve a better balance between representation storage and representation effect, $ss$ and $ts$ can be set in combination with the practical use when employing this model to generate human-like representations.

In this research, we present three strategies for cooperation inside the same computational model.
Here, the relationship between the the cooperation strategies and two types of representations' differences will be discussed.
Firstly, the ratio of the best results obtained in similar concepts tests with identical parameters is counted.
Then we consider each concept representation dataset alone, and count the closeness to humans when no human-like concept cooperation is performed.
Next, calculate the closeness difference between each two types of concept representations.
Finally, the correlation coefficients between the best ratio array and the closeness difference array were calculated.

\begin{table}[h!]
\begin{center}
\caption{The Relationship Between the Best Ratios for the Concepts in Three Distinct Cooperate Ways and the Closeness Differences}
\label{CorrelationBetweenDifferenceANDBestCOORatio}
\begin{tabular}{lccc}
\toprule	
Measure Datasets	& AND &OR &NOR  \\	
  \midrule
SimLex999	&-0.504353839	&0.680350214	&-0.692782248\\
MEN	&0.592628688	&-0.715378504	&-0.190924721\\
MTurk	&0.94203991	&-0.995577915	&0.990576611\\
\bottomrule
\end{tabular}
\end{center}
\end{table}

Table \ref{CorrelationBetweenDifferenceANDBestCOORatio} illustrates the relationship between the best ratios for the concepts in three distinct cooperate ways and the closeness differences between the two types of representations in the environment of the same evaluation dataset.
The findings demonstrate that the information differences between multisensory representations and text-derived representations are highly related to either cooperate way.
In other words, variations in how concepts with different attributes are represented using multisensory representations or text-derived representations can be closely tied to the most beneficial cooperate way.
However, there is no strong agreement regarding the direction of the relationship between closeness difference and the cooperate way, whether it is positive or negative.
Further research by computational neuroscience and cognitive psychology may be necessary to determine which precise factors determine the best cooperate way.

\section{Discussion}
In this study, we concentrate on the mechanisms by which humans acquire concepts and construct computational models that combine with information derived from text and multiple senses to generate representations.
With the support of this model, we could generate concept representations that are more similar to human cognition by overcoming the issue that the two types of representations have diverse sources and inconsistent dimensions.
By strictly adhering to the paradigm of mechanism to computational model, the research in this paper aims to facilitate the development of brain-inspired intelligence.

However, there are still a lot of difficulties that are worth exploring.
The model architecture in this research is based on previously published findings in cognitive psychology and computational neuroscience.
At the micro-scale, we do not find relevant evidence and can only explore it on the computational model, such as how exactly the two types of information of a concept merge and whether the two forms of information for different concepts are biased.
It is challenging to undertake more in-depth mechanistic studies, which is mostly because the linguistic abilities involved in concept learning are high-level cognitive functions.
We are eager to follow the development of relevant research to further refine our computational model.

When it comes to the two different types of concept representation datasets.
The currently available multisensory representation datasets of concepts are based on cognitive psychologists recruiting volunteers for labeling, which are highly interpretable, but the cost of labeling is more “expensive” and the scale of labeling is usually constrained.
In contrast, computational linguists collect text corpora and develop text-derived representations via machine learning methods, which are widely accessible but have a limited ability to be interpreted.
Since these two types of representations are inherently non-homogeneous, it is important to investigate if a mapping between the two types of representation datasets can be created by algorithmic design. 
This avoids the disadvantages of both while embracing the benefits of both, as well as increasing the scale of the datasets and supporting the development of concept learning models.

Additionally, the development of unsupervised spiking neural network algorithms, particularly the relationship between neural synchronization and information fusion equilibrium, deserves further investigation.
It can simulate human concept acquisition utilizing computational model provided in this research.
It is also strongly anticipated that AI agents will be able to use concept knowledge to produce downstream human-like actions in order to accomplish human-like cognitive tasks such as common sense knowledge reasoning, social cognition, and so on.

\section{Methods}
\begin{figure}[h!]
  \centering
  \includegraphics[width=1.0\textwidth]{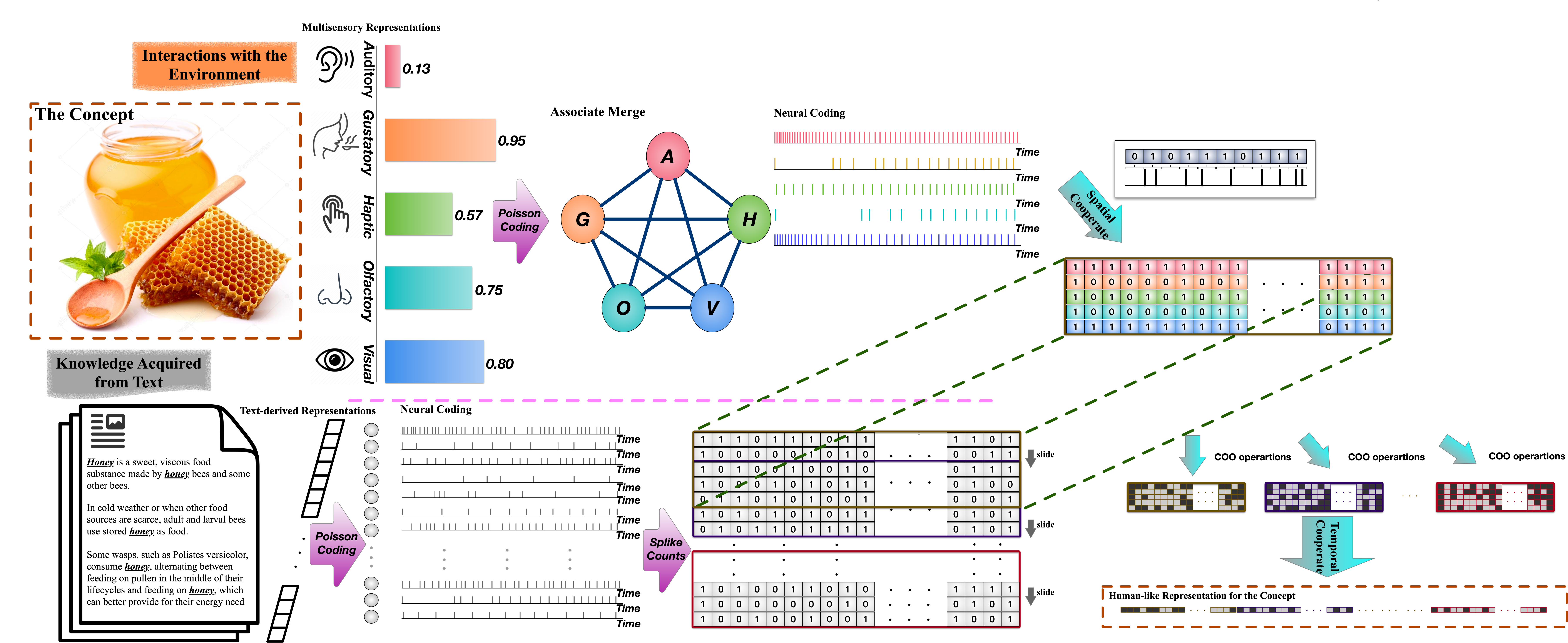}\\
  \caption{The Computational Model on Human-like Concept Learning}
  \label{HumanLikeTriFramework}
\end{figure}


We create a computational model for concept learning that properly mimics the architectural paradigm of human cognition.
The three parts of the computational model (multisensory information processing module, text-derived information processing module, and semantic control module) correspond one by one to the three networks of human cognition (multimodal experimential system, language-based semantic system, and semantic control system), as indicated in Figure \ref{Inspiration}.

Earlier research has shown that publicly available concepts' multisensory representations and text-derived representations datasets can mirror human cognition well\cite{wang2022statistical}.
Although there is no doubt that both types of representations serve as input to our model, there are two significant concerns to consider.
One is that the data originates from difference sources. 
The multisensory representations of concepts are based on emboded theory and labeled by cognitive psychologists, while the text-derived representations of concepts are based on distributed hypotheses and trained by computational linguists with language models and corpus data.
Another concern is that the data's dimensions vary tremendously.
The features of concepts' multisensory representations are closely tied to each modality's category.
The majority of these datasets have a common dimension of 5, which represents the perceptual potency of concepts learned through the five senses of vision, audio, smell, taste, and touch.
The dimension size of text-derived representations can be set according to the language model and the relevant corpus.
However, the dimensions that typically utilized are above 100, as with the 300-dimension version of the text-derived representations are frequently used.
Our computational model encounters a major challenge in resolving the dimension difference in order to preserve the balance of multisensory and text-derived representations and the coordination of both information.

In order to address the first issue, we consider the fact that even though learning new concepts, the human brain encounters a variety of external inputs, all of which take the form of unified signal transmission within the brain.
We transform the two types of concept representations into independent neural stimuli, and the original representation into spike train, thus unifying the data representation and mimicking the initial process of concept acquisition in the human brain.
To solve the second problem, we creatively exploit the properties of spiking neural networks.
The temporal domain information is unified by controlling the duration $T$ of neurons' performance that are simulated by external information, and the information between the two representations is integrated via sliding coordination.
The final human-like concept representation is obtained once the concept representation has been coordinated spatially and temporally.
Figure \ref{HumanLikeTriFramework} illustrates the framework diagram of the computational model for human-like concept learning, and the detailed work is discussed below.

\subsection{Poisson Coding}
Whether it is a text-derived representation or a multisensory representation of a concept, each dimensional value $r$ of the vector denotes a particular level of information intensity.
We expect using an encoding strategy such that the spike sequence produced based on $r$ is strongly associated and biologically interpretable with $r$ for a given duration $T$.
Assume that $M$ small intervals of length $\Delta t$ exist in $T$, that the lengths of these small intervals are so small that at most one spike can occur in them, and that the random variable determining whether a spike occurs in $\Delta t$ follows a Bernoulli distribution with expectation $\Delta t \cdot r$.
The probability of $n$ spikes firing in the period of time $T$ is then

\begin{align}
    P_T[n] &= lim_{\Delta t \rightarrow 0}\frac{M!}{(M-n)!n!}(r\Delta t)^n(1-r\Delta t)^{M-n}\\
        	&= lim_{\Delta t \rightarrow 0}\frac{M!}{(M-n)!n!}(r\Delta t)^n[(1-r\Delta t)^{(\frac{1}{-r\Delta t})}]^{-r\Delta t(M-n)}\\
    	      &\approx lim_{\Delta t \rightarrow 0} \frac{M^n}{n!} (r\Delta t)^n [(1-r\Delta t)]^{(\frac{1}{-r\Delta t})}]^{-r\Delta t M} \\
	      & =  lim_{\Delta t \rightarrow 0} \frac{(T/\Delta t)^n}{n!} (r\Delta t)^n [(1-r\Delta t)]^{(\frac{1}{-r\Delta t})}]^{-r\Delta t (T/\Delta t)}  \\
	      & = \frac{(rT)^n}{n!}exp(-rT)
\end{align}

This happens to obey the Poisson distribution, and this encoding is biologically, universal, as in the macaque visual cortex\cite{bair1994power}, medial temporal lobestudies\cite{softky1992cortical}.
In engineering, random numbers $x_{rand}$ with a uniform distribution are taken from the $[0,1]$ interval and a spike generated at that precise moment if $r\cdot \Delta t > x_{rand}$, otherwise not.
Additionally, we do min-max normalization preprocessing to obtain each dimensional information intensity $r$ before putting it into the model.

\subsection{Multisensory Information Processing Module}
The multisensory information processing module's primary responsibility in the computational model we created for human-like concept learning is to integrate and convert the various modal perceptual information of concepts into spike trains.
Wang \emph{et al.} previously developed a concept learning framework for multisensory information integration in this area\cite{wang2022multisensory}.
Associative merge(AM) and independent merge(IM) are the two distinct merge paradigms that were put forth under that framework.
AM presupposes that the information is related across modalities before integration, whereas IM assumes that the information is independent across modalities.
AM is an architecture in which the corresponding neurons of the various modalities are interconnected but not self-connected, while IM is a two-layer spiking neural network model.
Both types of frameworks have been shown to integrate information from various modalities (visual, auditory, tactile, gustatory, olfactory) well and to obtain integration representations with good results.
It has been demonstrated that both forms of paradigms effectively integrate data from a range of modalities, and produce integration representations that perform well.

This approach will be used in the multisensory information processing module to obtain the multisensory neural encoding for concepts.
Moreover, we'll apply the AM paradigm in this paper.
The module's input is a min-max normalized multisensory representation $\vec{m} = [m_A, m_G, m_H, m_O, m_V]$, where the subscripts $A, G, H, O, V$ standing for the five most prominent sensory inputs: auditory, gustatory, hatic, olfactory, and visual.
Based on the perceptual intensity and Poisson coding of each modality, the spike trains of each neuron are initially determined.
The correlation coefficients of the inter-modal data are used to estimate the connection weights between different modal neurons, i.e., $w_{i, j} = Corr(i, j)$, $i, j \in [A, G, H, O, V]$  and $Corr(i, j)$ stands for the relevance between each two sensory modalities.
With the leaky integrate-and-fire model, all neurons produce spikes.
The output of this module is the spike distribution of all neurons in the time interval $[0,T]$. 
If there is a spike in one time step, the cell in the spike distribution matrix $M^{spike}_{D_{ms}*T}$ is recorded as 1, otherwise it is 0.

\subsection{Text-derived Information Processing Module}

The text-derived information processing module of the our computational model is devoted to converting text-derived representations of concepts into spike trains, identical to the multisensory information processing module.
In this field, Wang \emph{et al.}\cite{wang2019biological} had converted dense word vectors into binarycodes after transforming text-derived representations into spike trains.
The Poisson coding based algorithm has proved that it not only enables the generated binary representations to maintain the original representations well, but also greatly reduces the storage space and performs well in NLP downstream tasks.
This method will also be utilized in this study to create the spike distribution matrix $T^{spike}_{D _{text}*T}$ from the normalized text-derived representation $\vec{t} = [t_{dim1}, t_{dim2}, t_{dim3}, \cdots, t_{dims(D_{text} -1)}, t_{dims(D_{text})}]$, which is also based on Poisson coding.
It is remarkable that we may regulate the spike distribution matrices' ($M^{spike}_{D_{ms}*T}$ and $T^{spike}_{D _{text}*T}$) length by adjusting the spike trains' temporal width $T$, thus unifying the two different representations of concepts in the temporal domain.

\subsection{Semantic Cooperate Module}
The two modules mentioned above facilitate the unifying of the signal forms as spike trains and fulfill the spike representation of the two different types of representations for the same concept.
The characteristics of spiking neural networks also empower the dimensional unification in temporal.
The collaboration of these two types of information will be accomplished in the semantic control module through the operation of spatial cooperate and temporal cooperate, and finally the human-like concept representation will be generated.

\subsubsection{Spatial Cooperate}
By setting the same spike train recording time $T$, we can make the width of the spike distribution matrix same.
Nevertheless, the height of the spike distribution matrices $M^{spike}_{D_{ms}*T}$ and $T^{spike}_{D _{text}*T}$ obtained by the multisensory information processing module and the text-derived information processing module are still different.
This is primarily due to the fact that the dimensions of the two types of concept representations differ by dozens of times.
To overcome this challenge, we use a sliding window to extract the block $T^{(i)}_{block}$ that is identical to the time-space domain of the multisensory representation spike distribution matrix, where the spatial stride is denoted by $ss$.

\begin{equation}
T^{(i)}_{block}=T^{spike}_{D_{ms}*T}=T^{spike}_{D_{text}*T}[i*ss:i*ss+D_{ms}, :]
\end{equation}

The matrices $T^{(i)}_{block}$ and $M^{spike}_{D_{ms}*T}$ are the same size.
As follows, we'll carry out the spatial cooperate operation in binary space.

\begin{equation}
SC^{(i)} = \mathcal{S}(T^{(i)}_{block}, M^{spike}_{D_{ms}*T})
\end{equation}
where $\mathcal{S}(\cdot)$ represents the binary operation for $T^{(i)}_{block}$ and $M^{spike}_{D_{ms}*T}$, specifically, three operations such as “AND”, “OR” and “NOR” are set, where

\begin{equation}
\mathcal{S}^{AND}(A, B) = C, \ where \ C[j, k] = \begin{cases} 
1 &A[j, k]=B[j, k]=1\\
0 &otherwise
\end{cases}
\end{equation}

\begin{equation}
\mathcal{S}^{OR}(A, B) = C, \ where \ C[j, k] = \begin{cases} 
0 &A[j, k]=B[j, k]=0\\
1 &otherwise
\end{cases}
\end{equation}

\begin{equation}
\mathcal{S}^{NOR}(A, B) = C, \ where \ C[j, k] = \begin{cases} 
0 &A[j, k]=B[j, k]\\
1 &otherwise
\end{cases}
\end{equation}

\subsubsection{Temporal Cooperate}
Considering that the two types of information overlap due to the spatial cooperation mentioned above, it not only makes the information cooperation more thorough but also results in some information redundancy.
Therefore, in order to improve the final representation's noise immunity and characterisation capability, we construct the temporal cooperate operation to reduce the information of $SC^{(i)}$ in the temporal space.
The temporal stride, denoted as $ts$, is set under each block to execute the temporal domain information cooperation in order to achieve the temporal cooperation between multisensory representations and text-derived representations. 
$TC^{(i)} = \mathcal{T}(SC^{(i)})$.

where $\mathcal{T}(\cdot)$ denotes the binary matrix $A_{J*K}$ after performing spatial coordination, as follows.
\begin{equation}
\begin{aligned}
\vec{a} &= A[1,:] \oplus A[2,:] \oplus A[3,:] \oplus \cdots \oplus A[D_{ms}-1,:] \oplus A[D_{ms},:] \\
\mathcal{T}(\vec{a}) &= \Gamma(\vec{a}[0,ts]) \oplus \Gamma(\vec{a}[ts, 2*ts]) \oplus,  \cdots, \oplus \Gamma(\vec{a}[ \lfloor \frac{J*K}{ts} \rfloor*ts, J*K]) 
\end{aligned}
\end{equation}
and $\Gamma(\cdot)$ is an indicator function, which means if 1 in the interval the value is 1, otherwise the value 0.
Eventually, the information from each block is concatenated together and the generation of human-like concept representations has been achieved.

\begin{equation}
C_{HumanLike} = \oplus_{i} TC^{(i)}
\end{equation}


\bibliography{ref}
\end{document}